\documentclass[pmlr,twocolumn,10pt]{jmlr}

\usepackage{booktabs}
\usepackage{siunitx}
\usepackage[switch]{lineno}
\usepackage{xcolor}
\usepackage{colortbl}
\usepackage{float}
\usepackage{algorithm}
\usepackage{algorithmic}
\usepackage{verbatim}
\usepackage{multirow}

\newcommand{\crisp}{\textsc{CRISP}}
\newcommand{\dataset}{\textsc{CRITICAL}}

\definecolor{headerblue}{RGB}{70,130,180}

\title[CRISP: End-to-End Processing of Large-scale Multi-institutional OMOP CDM Data]{\large The CRITICAL Records Integrated Standardization Pipeline (CRISP):\ End-to-End Processing of Large-scale Multi-institutional OMOP CDM Data}

\author{
    \Name{Xiaolong Luo} \Email{xiaolongluo@fas.harvard.edu} \\
    \addr School of Engineering and Applied Sciences, Harvard University, Cambridge, MA 02138, USA
    \AND
    \Name{Michael Lingzhi Li} \Email{mili@hbs.edu} \\
    \addr Harvard Business School, Harvard University, Boston, MA 02163, USA
}

\begin{document}

\maketitle

\begin{abstract}
While existing critical care EHR datasets such as MIMIC and eICU have enabled significant advances in clinical AI research, the CRITICAL dataset opens new frontiers by providing extensive scale and diversity---containing 1.95 billion records from 371,365 patients across four geographically diverse CTSA institutions. CRITICAL's unique strength lies in capturing full-spectrum patient journeys, including pre-ICU, ICU, and post-ICU encounters across both inpatient and outpatient settings. This multi-institutional, longitudinal perspective creates transformative opportunities for developing generalizable predictive models and advancing health equity research. However, the richness of this multi-site resource introduces substantial complexity in data harmonization, with heterogeneous collection practices and diverse vocabulary usage patterns requiring sophisticated preprocessing approaches.

We present CRISP (CRITICAL Records Integrated Standardization Pipeline) to unlock the full potential of this valuable resource. CRISP systematically transforms raw Observational Medical Outcomes Partnership Common Data Model data into ML-ready datasets through: (1) transparent data quality management with comprehensive audit trails, (2) cross-vocabulary mapping of heterogeneous medical terminologies to unified SNOMED-CT standards, with deduplication and unit standardization, (3) modular architecture with parallel optimization enabling complete dataset processing in $<$1 day even on standard computing hardware, and (4) comprehensive baseline model benchmarks spanning multiple clinical prediction tasks to establish reproducible performance standards. By providing processing pipeline, baseline implementations, and detailed transformation documentation, CRISP saves researchers months of preprocessing effort and democratizes access to large-scale multi-institutional critical care data, enabling them to focus on advancing clinical AI. The complete source code, baseline models, and documentation are publicly available.
\end{abstract}

\begin{keywords}
Electronic Health Records, CRITICAL Dataset, Intensive Care Unit, Multi-institutional Data, Critical Care, Data Processing Pipeline, Healthcare AI, Real-world Data
\end{keywords}

\paragraph*{Data and Code Availability}
The CRITICAL dataset is available under a data use agreement at \url{https://critical.fsm.northwestern.edu}. CRISP source code, documentation, and processing scripts are publicly available at \url{https://github.com/AaronLuo00/CRISP-Pipeline}.


\paragraph*{Institutional Review Board (IRB)}
This research has been designed by Harvard IRB as Not Human Subject Research. IRB Protocol information will be provided if the paper is accepted.

\section{Introduction}

The rapid advancement of artificial intelligence (AI) has revolutionized healthcare through its integration with electronic health records (EHRs), enabling unprecedented capabilities in clinical prediction, diagnosis, and decision support. Recent breakthroughs have demonstrated the ability to accurately predict multiple medical events from EHR data \citep{rajkomar2018scalable,jiang2023health, grout2024predicting, hegselmann2025large}, with models achieving performance comparable to clinical experts in various domains including mortality prediction, disease diagnosis, and treatment recommendation. These successes have increased interest in developing AI systems that can assist clinicians in real-time decision-making \citep{tomasev2019clinically}, predict patient deterioration hours before clinical manifestation \citep{lauritsen2020explainable}, and optimize resource allocation in increasingly strained critical care settings where timely intervention can significantly impact patient outcomes \citep{komorowski2018artificial,gutierrez2020artificial}.

However, the promise of AI in healthcare critically depends on access to large-scale, diverse, and well-structured clinical data, a fundamental challenge that limit the development of truly generalizable AI models \citep{futoma2020myth,kelly2019key}. Most existing models are trained on single-institution datasets, raising concerns about their transferability across different healthcare systems, patient populations, and clinical practice patterns \citep{zech2018variable,nestor2019feature}. Furthermore, the heterogeneity in data collection practices, vocabulary usage, and documentation standards across institutions creates a substantial barrier to developing robust, multi-institutional AI systems that can benefit diverse patient populations \citep{gianfrancesco2018potential,obermeyer2019dissecting}. This challenge is particularly acute in critical care settings, where the complexity of patient conditions, the diversity of monitoring equipment, and the urgency of clinical decisions demand models that can generalize across varying institutional protocols and patient demographics \citep{sendak2020machine,shah2019artificial}.

\subsection{The Data Challenge in Healthcare AI}

Pioneering datasets like MIMIC-III and MIMIC-IV have established the foundation for critical care AI research, with MIMIC-IV comprising over 65,000 intensive care unit (ICU) patients and more than 200,000 emergency department (ED) patients, enabling groundbreaking advances in mortality prediction, treatment optimization, and clinical decision support \citep{johnson2016mimic,johnson2023mimic}. The eICU Collaborative Research Database further expanded the field by demonstrating the value of multi-center data ($\sim$200,000 ICU admissions across 208 hospitals), pioneering cross-institutional research approaches \citep{pollard2018eicu}. These foundational resources have trained a generation of researchers and established methodological standards that continue to guide the field.

Building upon these essential contributions, the CRITICAL dataset extends the research landscape by providing 1.95 billion records from 371,365 patients across four Clinical and Translational Science Awards (CTSA) sites, offering complementary strengths including full-spectrum patient journeys (pre-ICU, ICU, and post-ICU), extended longitudinal tracking, and diverse geographic representation \citep{critical_dataset_2025}. While this scale and diversity enable more generalizable modeling, they also introduce cross-site semantic heterogeneity—differences in vocabulary usage, coding practices, units, and temporal granularity—demanding transparent, reproducible data preprocessing for cleaning, standardization, and harmonization. CRISP addresses these challenges by providing a modular, reusable pipeline that not only accelerates research but also ensures consistency with the methodological standards established by the MIMIC and eICU communities.

\subsection{Our Contributions}

To address these critical needs for standardization and unified data formats, we present CRISP (CRITICAL Records Integrated Standardization Pipeline), a comprehensive solution that transforms CRITICAL's 1.95 billion raw records into ML-ready formats. Our contributions include:

\textbf{(1) Five-stage preprocessing pipeline} that systematically transforms raw Observational Medical Outcomes Partnership Common Data Model (OMOP CDM) data through exploratory analysis, data cleaning, vocabulary mapping, standardization, and ICU cohort extraction with comprehensive audit trails ensuring reproducibility.

\textbf{(2) Scalable parallel architecture} that processes the entire 278.97 GB dataset in under 24 hours using 12 CPU cores and 64GB RAM through optimized chunked processing and parallel optimization, making large-scale multi-institutional data processing accessible to resource-constrained research teams.

\textbf{(3) Comprehensive benchmarks} across four critical prediction tasks using multiple model architectures, establishing reproducible baselines for the research community.

\textbf{(4) Open-source implementation} with complete code, documentation, and processed datasets, saving researchers months of preprocessing effort.


\section{Related Work}

\subsection{Clinical Data Processing Pipelines}

As clinical datasets grow in scale and complexity, the need for standardized, reusable pipelines becomes increasingly critical. In particular, there has been many processing pipelines designed for MIMIC-III, MIMIC-IV, eICU and other EHR datasets. Notable contributions include MIMIC-Extract \citep{wang2020mimic} for cohort extraction, multitask benchmarks \citep{harutyunyan2019multitask} establishing standard prediction tasks, COP-E-CAT \citep{mandyam2021cop} for modular preprocessing, an extensive MIMIC-IV pipeline \citep{gupta2022extensive}, METRE \citep{liao2023multidatabase} for cross-database validation, and reproducibility MIMIC benchmark \citep{purushotham2018benchmarking}.

These pipelines have established a solid foundation for processing single-institution EHR datasets and have achieved remarkable success within their respective domains. However, extending them to multi-institutional environments poses several challenges. Existing pipelines are typically optimized for dataset-specific schemas (e.g., MIMIC’s custom structure or eICU’s format). Additionally, they often assume a single-vocabulary system, whereas multi-institutional environments incorporate multiple overlapping vocabularies requiring sophisticated cross-vocabulary harmonization. Finally, many of these pipelines utilize single-threaded architectures that are sufficient for moderate-scale datasets, but inadequate for multi-site, billion-row CDM tables.

\crisp{} builds upon these prior efforts by introducing a parallelized pipeline tailored specifically for a multi-institutional setting. It incorporates systematic data cleaning, schema standardization, and multi-vocabulary harmonization to enable large-scale, multi-institutional processing. Our implementation achieves a 4–6$\times$ speedup compared to serial execution and completes full-dataset processing in approximately 20 hours on commodity hardware. We also release reproducible benchmarks covering both traditional machine learning models and deep learning architectures.


\subsection{Broader Context and Challenges}

\textbf{Multi-institutional Harmonization and Standardization:}  
The heterogeneity of medical vocabularies across institutions creates fundamental challenges for multi-site data integration. \citet{henke2024conceptual} proposed systematic harmonization approaches to address schema heterogeneity across OMOP implementations, identifying multiple sources of incompatibility requiring comprehensive harmonization processes. \citet{wang2025scoping} examined OMOP CDM adoption challenges in specialized domains like oncology, revealing significant gaps in cancer-specific concept coverage. This vocabulary heterogeneity leads to severe feature matrix sparsity: when identical clinical concepts are encoded differently across sites, each feature is populated only by a subset of institutions. The resulting matrices are dominated by missing values, introducing noise and degrading model performance \citep{che2018recurrent}. As demonstrated in Figure \ref{fig:vocab_distribution}, this vocabulary heterogeneity is particularly pronounced in multi-institutional datasets. \crisp{} addresses this challenge by systematically analyzing vocabulary distributions in the CRITICAL dataset, constructing cross-vocabulary mappings to unified Systematized Nomenclature of Medicine Clinical Terms (SNOMED-CT) \footnote{hereafter refer as SNOMED} standards, and standardizing different units to Unified Code for Units of Measure (UCUM) specifications \citep{schadow2009unified}, thereby consolidating fragmented features into dense, semantically consistent representations essential for robust multi-institutional EHR model training.


\textbf{Clinical Benchmarks:} Foundational work established standard prediction tasks (e.g., mortality, length-of-stay) and demonstrated that preprocessing strongly affects performance \citep{johnson2017reproducibility,rocheteau2021temporal}. Building on this, we release reproducible benchmarks over \dataset{} using \crisp{}-processed OMOP data, enabling fair comparisons across models.

\begin{figure*}[!h]
\centering
\includegraphics[width=0.85\textwidth]{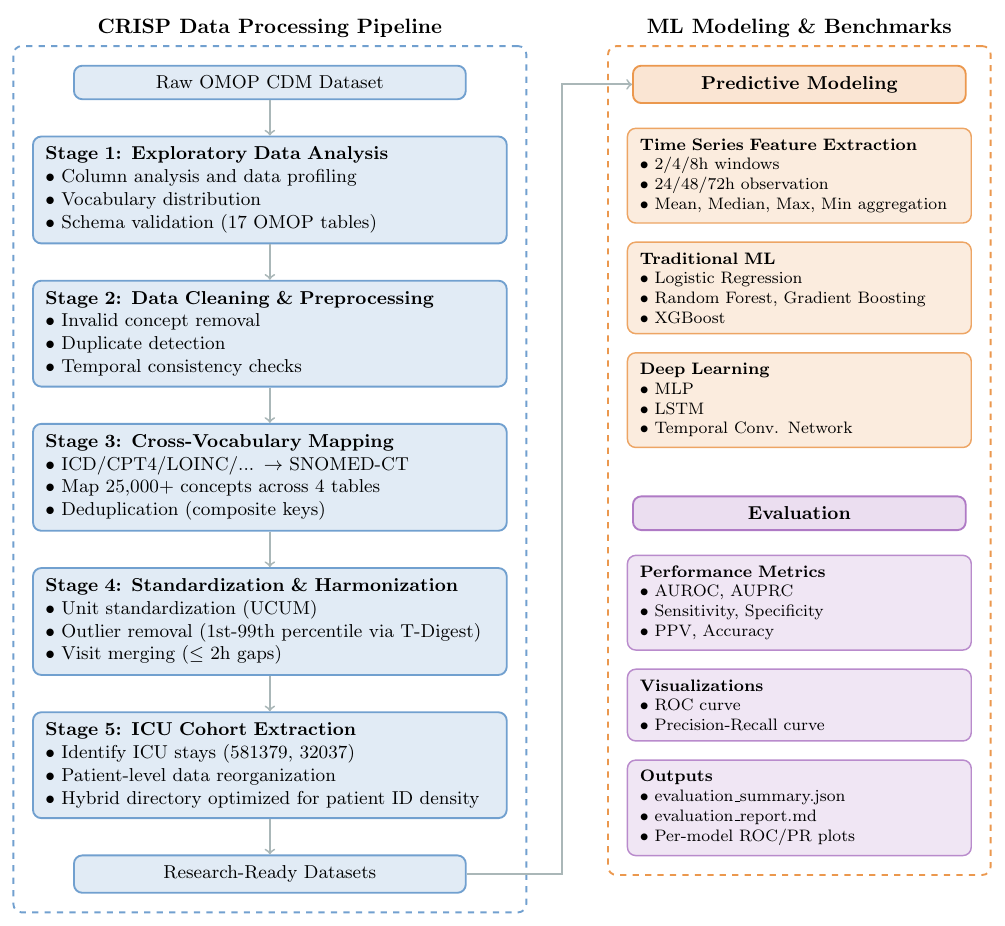}
\caption{CRISP Pipeline Architecture: Five-stage data processing pipeline for the CRITICAL dataset.}
\label{fig:pipeline}
\end{figure*}

\section{The CRITICAL Dataset}

The CRITICAL dataset represents the first cross-CTSA initiative to create a multi-site, multi-modal, de-identified clinical dataset combining both deep longitudinal coverage and broad institutional diversity. Developed collaboratively across four CTSA sites (Northwestern, Tufts, Washington University in St. Louis, and University of Alabama at Birmingham), CRITICAL encompasses 1.95 billion records from 371,365 patients \citep{critical_dataset_2025}, establishing it as the largest publicly shared, disease-independent benchmarking dataset for critical care research. Built on OMOP CDM v5.3, the repository spans 17 tables totaling 278.97 GB (Table~\ref{tab:appendix_volumes}), with MEASUREMENT alone containing 1.4 billion rows. The dataset includes 38 million visits and 28 million unit-level records, averaging 5,242 rows per patient across all tables.

CRITICAL provides comprehensive patient care journeys with a median observation period of 3.11 years and maximum spanning 31.8 years (Table~\ref{tab:appendix_temporal}). This extensive temporal coverage captures pre-ICU, ICU, and post-ICU encounters across both inpatient and outpatient settings, with patients averaging 102.3 visits throughout their observation periods. This multi-institutional, longitudinal perspective introduces substantial vocabulary heterogeneity—150,671 unique source concepts across 30 vocabularies\footnote{After deduplication across related tables, the dataset contains over 110,000 unique clinical concepts.}, with SNOMED alone accounting for 58.0\% (87,453 concepts). Figure \ref{fig:vocab_distribution} illustrates this vocabulary heterogeneity across major tables within the dataset (see Appendix~\ref{apd:second} for detailed distribution analysis), requiring systematic harmonization to unified standards.


This combination of extensive scale, institutional diversity, and longitudinal depth improves access to large-scale clinical data for AI research.

\section{Data Pipeline Overview}

\subsection{Pipeline Architecture}

To harness the CRITICAL dataset's scale and multi-institutional diversity described in Section 3, \crisp{} employs a five-stage processing framework that transforms the raw OMOP CDM tables into ML-ready dataset. The pipeline architecture consists of two primary components: (1) a core data processing module executing sequential stages of exploratory analysis, data cleaning, cross-vocabulary mapping, standardization, and patient data extraction with label generation; and (2) an optional predictive modeling module offering baseline implementations and evaluation benchmarks. This modular design allows researchers to utilize individual stages independently, customize processing parameters, or extend the pipeline with task-specific modification. (Figure \ref{fig:pipeline}).

The implementation leverages parallel processing strategies across all computationally intensive operations. Through chunked data loading and concurrent table processing, the pipeline handles billion-row tables within memory constraints while maintaining processing efficiency. Every transformation generates detailed audit trails—tracking removed records, vocabulary mappings, unit conversions, and outlier statistics—enabling complete reproducibility. This architecture processes the entire 278.97 GB CRITICAL dataset in under 24 hours using standard computational resources (12 CPU cores, 64GB RAM).

\subsection{Five-Stage Processing Pipeline}

\textbf{Stage 1: Exploratory Data Analysis.} This stage generates comprehensive dataset statistics that guide subsequent processing modules and provide researchers with detailed data understanding. The pipeline analyzes all 17 OMOP tables, producing: (1) column-level missingness analysis, automatically flagging columns with $>$95\% missing values for removal to simplify downstream processing; (2) table-level summaries including row counts (1.95 billion total), unique patient counts , memory usage, and temporal coverage (date ranges for each table); and (3) population-level statistics such as ICU admission rates, mortality rates , gender distributions, and age ranges. All statistics are exported as structured JSON files that subsequent modules consume to parameterize their operations.


\textbf{Stage 2: Data Cleaning and Preprocessing.} Building on Stage 1's column analysis, this stage systematically cleans 14 tables\footnote{The three tables not processed are LOCATION, CARE\_SITE, and PROVIDER, as they either lack meaningful information or are independent of patient-level data.} by addressing data quality issues. The pipeline performs three parallel cleaning operations: (1) \textit{invalid concept removal} filters out records where the primary concept ID field (e.g., measurement\_concept\_id, procedure\_concept\_id) is null, empty, or zero—these represent unmapped or invalid clinical codes that cannot be interpreted, eliminating approximately 2-5\% of records; (2) \textit{duplicate elimination} identifies and removes redundant records using composite keys (person\_id + concept\_id + datetime), preventing the same clinical event from being counted multiple times; and (3) \textit{temporal validation} ensures chronological consistency by verifying start times precede end times, removing records with future dates or impossible sequences. All removed records are archived in structured directories with detailed logs, enabling quality assessment and potential recovery.

\begin{figure}[!h]
\centering
  \includegraphics[width=1.1\columnwidth]{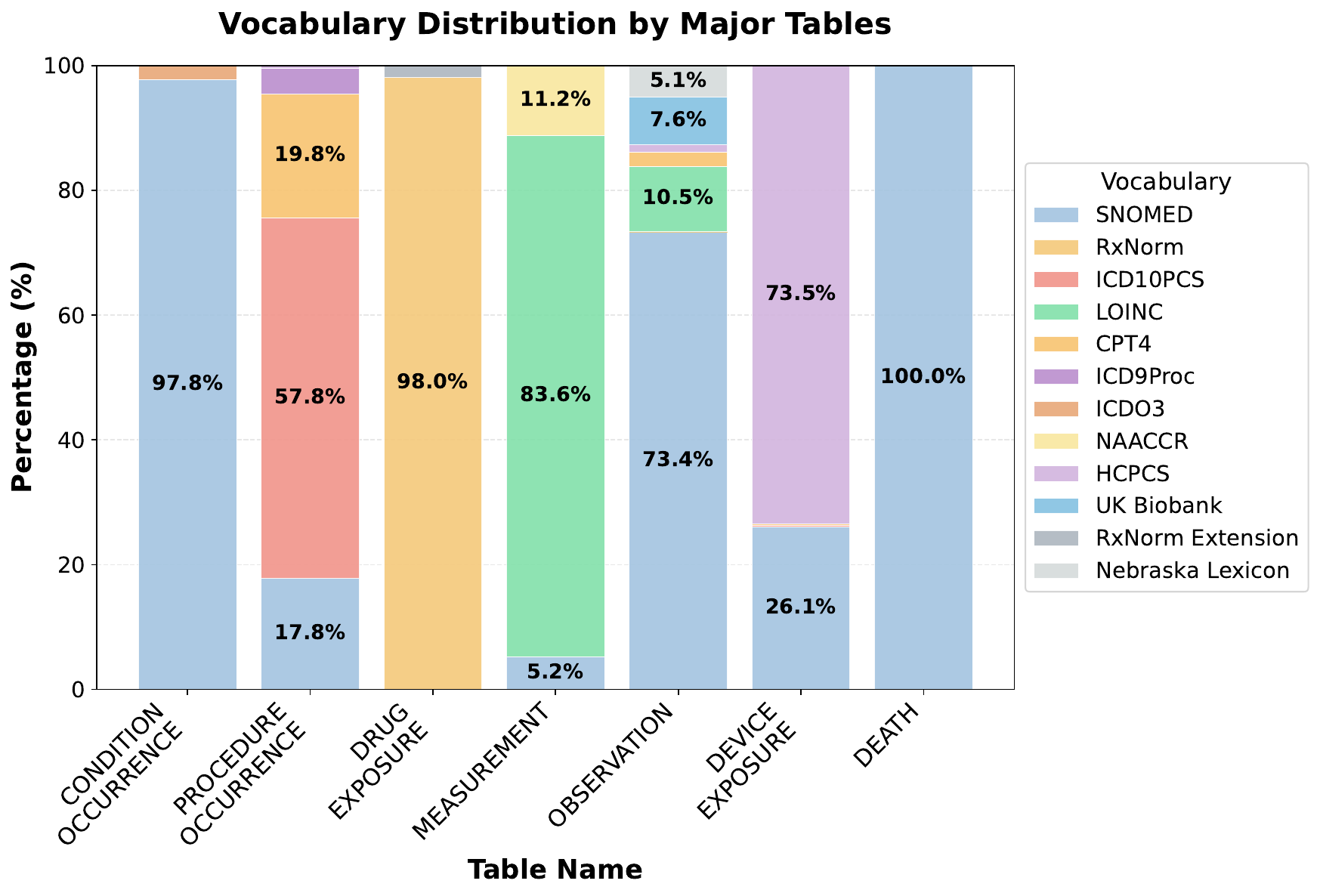}
\caption{Vocabulary distribution across major OMOP tables demonstrates.}
\label{fig:vocab_distribution}
\end{figure}

\begin{figure}[!h]
\centering
  \includegraphics[width=1.015\columnwidth]{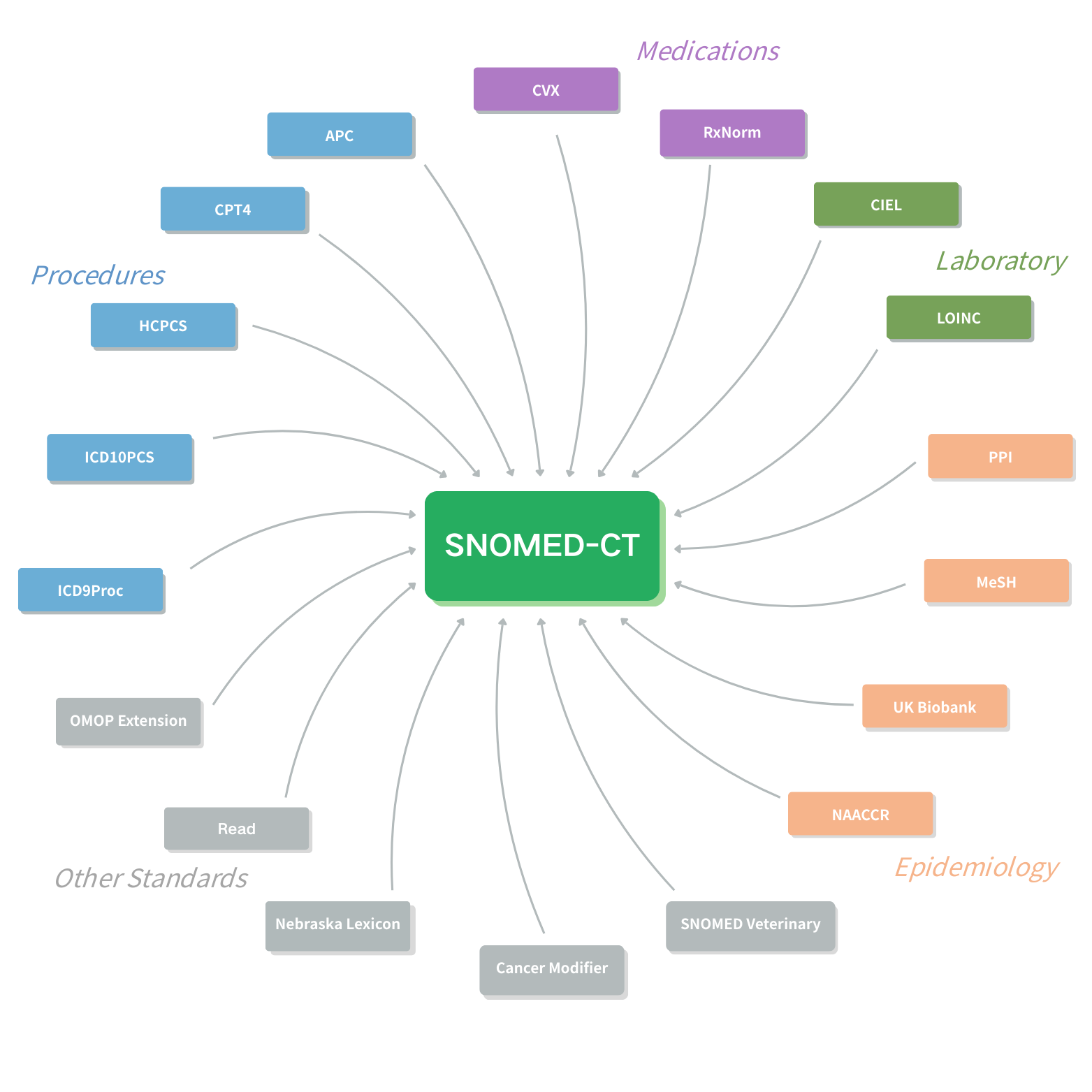}
\caption{Vocabulary harmonization: mapping diverse medical terminologies to unified SNOMED standards.}
\label{fig:vocab_mapping}
\end{figure}

\textbf{Stage 3: Cross-Vocabulary Concept Mapping.} This stage addresses the fundamental challenge of vocabulary heterogeneity in multi-institutional datasets, where different sites may use entirely different coding systems for the same clinical concepts. SNOMED was selected as the target vocabulary because it already represents the majority of concepts both globally (58.0\% of all concepts) and within most individual tables (Figure \ref{fig:vocab_distribution}), minimizing the required mapping effort while maximizing coverage. The pipeline targets four key tables\footnote{MEASUREMENT, OBSERVATION, PROCEDURE\_OCCURRENCE, DEVICE\_EXPOSURE} that exhibit the most complex vocabulary heterogeneity—for instance, PROCEDURE\_OCCURRENCE contains concepts from ICD10PCS (57.8\%), CPT4 (19.8\%), and SNOMED (15.4\%) as shown in Figure \ref{fig:vocab_distribution}. These tables are critical for clinical prediction yet suffer from severe fragmentation without harmonization. The pipeline applies pre-computed crosswalks to map diverse source vocabularies (including ICD9CM, ICD10CM, ICD10PCS, CPT4, LOINC, HCPCS, RxNorm, and others shown in Figure \ref{fig:vocab_mapping}) to unified SNOMED codes, harmonizing over 25,000 unique source concepts. Post-mapping deduplication removes redundancies from many-to-one mappings using composite keys (person\_id + SNOMED\_id + datetime), consolidating multiple source representations of the same clinical concept. 


\begin{figure}[!h]
\centering
\includegraphics[width=0.9\columnwidth]{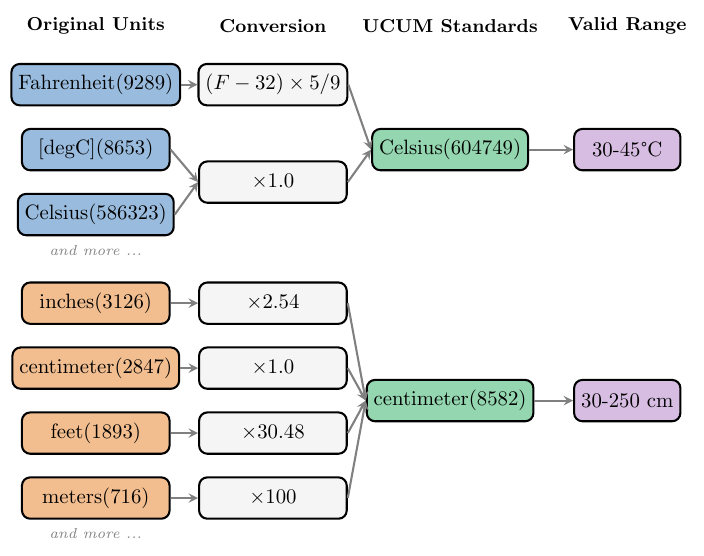}
\caption{Unit standardization pipeline: converting diverse measurement units to UCUM standards with outlier filtering.}
\label{fig:unit_standardization}
\end{figure}

\textbf{Stage 4: Data Standardization and Unit Harmonization.} This stage ensures measurement consistency and temporal coherence across the harmonized dataset. Processing tables from Stage 3's output, the pipeline performs four key standardization operations. First, \textit{outlier removal} applies T-Digest algorithms \citep{dunning2019computing} specifically to MEASUREMENT tables, computing memory-efficient percentiles across 1.4 billion records and filtering out outliers beyond the 1st and 99th percentiles\footnote{Configurable parameters, following the approach of \citet{gupta2022extensive}.}. Second, \textit{unit standardization} converts heterogeneous measurement units to UCUM standards—for example, temperature from Fahrenheit to Celsius, weight from pounds to kilograms, and height from inches to centimeters—while removing physiologically implausible values (illustrated in Figure \ref{fig:unit_standardization}). Third, \textit{visit consolidation} merges fragmented VISIT\_DETAIL and VISIT\_OCCURRENCE records within 2-hour windows, reconstructing continuous care episodes from 66 million visit records. Fourth, \textit{data type standardization} ensures consistent representation—NaN for missing values, integers for IDs, floats for measurements, and ISO format for datetimes. After Stage 4 completion, all critical tables have unified concepts, with missing, erroneous, and implausible values removed, and data formats standardized for machine learning readiness. 

\textbf{Stage 5: Patient Data Extraction and Label Generation.} This final stage transforms table-centric OMOP data into patient-centric structures required for machine learning. This stage performs three key operations. First, \textit{patient-level aggregation} reorganizes 1.95 billion records into 371,365 individual patient directories, consolidating each patient's complete medical history. Second, \textit{ICU cohort identification} scans VISIT\_DETAIL tables for ICU-specific concept IDs\footnote{(Using concept ID: 581379 for ICU stays, 32037 for critical care)}, and generating temporal marks for different patient stages (pre-ICU, during-ICU, post-ICU). The extraction employs parallel chunked processing to efficiently handle billion-scale data while maintaining patient-level integrity. 

\begin{figure}[!h]
\centering
\includegraphics[width=1.0\columnwidth]{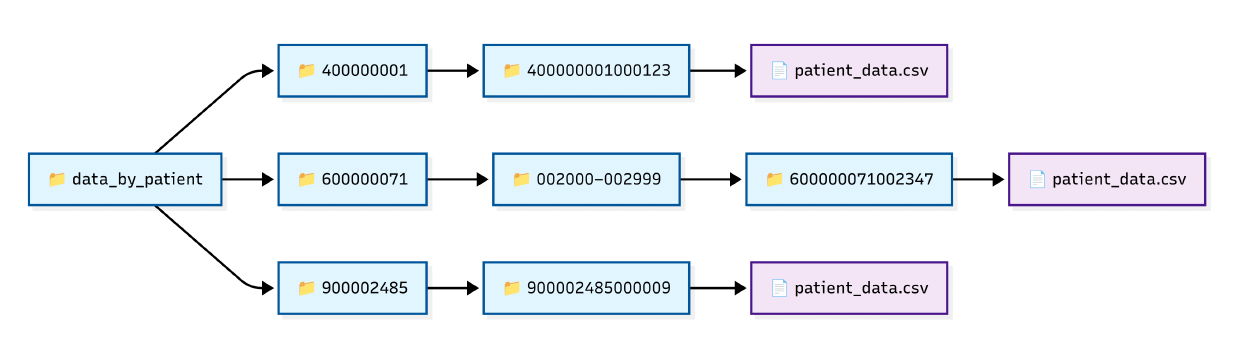}
\caption{Hybrid directory structure adapting to patient ID density for optimal file system performance.}
\label{fig:directory_structure}
\end{figure}

The resulting patient-indexed structure is organized using a hybrid directory system (Figure \ref{fig:directory_structure}) that adapts to the uneven distribution of 371,365 patient IDs. Low-density prefixes ($<$30,000 patients) use direct folder organization, while high-density prefixes like 600000071 ($>$30,000 patients) employ sub-directory layering (e.g., 002000-002999) to prevent file system degradation. This adaptive structure ensures efficient I/O performance while enabling direct patient queries for diverse ML tasks.

\subsection{Computational Performance Analysis}

CRISP leverages parallel processing across all five pipeline stages to handle large-scale clinical data efficiently. Each stage employs configurable worker pools (default 8 workers) for concurrent processing. This comprehensive parallelization achieves 4-6$\times$ speedup compared to sequential processing, reducing total pipeline execution time from approximately 4 days to $\sim$20 hours on standard hardware (12-core CPU, 64GB RAM). The substantial  acceleration, combined with memory-efficient strategies like T-Digest for percentile computation, chunked I/O operations, and the hybrid directory structure design, makes large-scale multi-institutional data processing feasible for resource-constrained research teams.

\section{Benchmark Tasks and Models}

To validate CRISP's effectiveness and establish reproducible baselines, we conduct comprehensive benchmark experiments on the harmonized CRITICAL dataset. Our evaluation framework tests both traditional and deep learning models across multiple clinical prediction tasks, revealing current performance limitations and opportunities for methodological advancement.

\subsection{Experimental Setup and Methodology}

Following MIMIC-Extract \citep{wang2020mimic}, we select the 800 most frequent clinical concepts from the harmonized dataset, extracting features from five key tables (MEASUREMENT, OBSERVATION, DRUG\_EXPOSURE, CONDITION\_OCCURRENCE, PROCEDURE\_OCCURRENCE). The observation window spans the first 24 hours of ICU admission, discretized into 4-hour bins to capture temporal dynamics. We evaluate seven model architectures spanning traditional ML (Logistic Regression, Random Forest, Gradient Boosting, XGBoost) and deep learning approaches (Multi-Layer Perceptron (MLP), Long Short-Term Memory networks (LSTM), Temporal Convolutional Networks (TCN)). All models employ 5-fold cross-validation, with the final results reported using 80\% of data for training and 20\% for testing. Detailed LSTM and TCN model architectures are provided in Appendix~\ref{apd:architectures}.


\subsection{Clinical Prediction Tasks}

Inspired by \cite{gupta2022extensive}, we define four binary classification tasks following standard ICU prediction benchmarks with varying time horizons:

\textbf{Mortality Prediction:} 7-day and 30-day in-hospital mortality using the first 48 hours of ICU data. This task addresses critical care triage and resource allocation decisions.

\textbf{Length of Stay:} Predicting ICU stays exceeding 3 and 7 days using the first 24 hours, essential for capacity planning and early intervention strategies.

\textbf{Readmission Risk:} 7-day, 30-day, and 90-day readmission prediction using the last 48 hours before discharge, crucial for discharge planning and follow-up care coordination.

\textbf{Sepsis Onset:} Detecting sepsis development after ICU admission, within 48 hours, and within 7 days, enabling timely antibiotic therapy and aggressive treatment protocols.

All tasks maintain 48-hour gaps between observation and prediction windows to prevent label leakage. We select 800 high-frequency features: 400 from MEASUREMENT, 200 from OBSERVATION, 100 from DRUG\_EXPOSURE, and 50 each from CONDITION\_OCCURRENCE and PROCEDURE\_OCCURRENCE, balancing computational  efficiency with clinical coverage.

\begin{table*}[!h]
\centering
\caption{Clinical Prediction Performance (AUROC)}
\label{tab:comprehensive}
\footnotesize
\begin{tabular}{llccccccc}
\toprule
\textbf{Task Category} & \textbf{Prediction Target} & \textbf{LR} & \textbf{RF} & \textbf{GB} & \textbf{XGB} & \textbf{MLP} & \textbf{LSTM} & \textbf{TCN} \\
\midrule
\multirow{2}{*}{\textbf{Mortality}} 
& 7-day & 0.684 & 0.676 & 0.763 & 0.781 & 0.814 & 0.697 & 0.705 \\
& 30-day & 0.708 & 0.732 & 0.802 & 0.804 & 0.835 & 0.764 & 0.778 \\
\midrule
\multirow{2}{*}{\textbf{Length of Stay}} 
& LOS $>$ 3 days & 0.651 & 0.702 & 0.732 & 0.735 & 0.756 & 0.689 & 0.711 \\
& LOS $>$ 7 days & 0.666 & 0.702 & 0.746 & 0.748 & 0.767 & 0.687 & 0.719 \\
\midrule
\multirow{3}{*}{\textbf{Readmission}} 
& 7-day & 0.641 & 0.703 & 0.739 & 0.755 & 0.748 & 0.663 & 0.681 \\
& 30-day & 0.619 & 0.699 & 0.726 & 0.735 & 0.743 & 0.652 & 0.698 \\
& 90-day & 0.635 & 0.695 & 0.741 & 0.746 & 0.737 & 0.663 & 0.702 \\
\midrule
\multirow{3}{*}{\textbf{In ICU Sepsis}} 
& After ICU & 0.879 & 0.895 & 0.897 & 0.898 & 0.883 & 0.884 & 0.871 \\
& Within 48h & 0.836 & 0.870 & 0.883 & 0.882 & 0.912 & 0.799 & 0.845 \\
& Within 7 days & 0.904 & 0.899 & 0.904 & 0.908 & 0.902 & 0.852 & 0.876 \\
\bottomrule
\end{tabular}
\end{table*}

The comprehensive results on the full CRITICAL dataset (Table \ref{tab:comprehensive}) reveal significant room for improvement in multi-institutional clinical prediction. Even with CRISP's harmonization and sparsity reduction, the best models achieve only 0.619-0.755 AUROC for readmission tasks, highlighting the inherent complexity of predicting patient trajectories across heterogeneous institutions. This performance gap motivates further research into advanced feature selection and utilization strategies, novel model architectures, as well as leveraging the extensive pre-ICU information uniquely available in CRITICAL.

\section{Discussion and Conclusion}

\textbf{Core Contribution.} CRISP is the first end-to-end processing pipeline specifically designed for the large-scale multi-institutional CRITICAL dataset. The pipeline systematically addresses the CRITICAL dataset's complexity—150,671 unique concepts across 30 vocabularies and 1.95 billion records—through five integrated stages: (1) systematic exploratory data analysis (2) comprehensive cleaning that removes invalid concepts, duplicates, and temporal inconsistencies across 14 tables; (3) cross-vocabulary mapping that harmonizes four key tables to unified SNOMED standards; (4) data standardization with outlier removal, unit conversion, and visit merging; and (5) patient-centric extraction that generates ML-ready features. Through optimized parallel processing and chunking strategies, CRISP achieves 4-6× speedup over sequential approaches, processing the entire 278.97 GB dataset in approximately 20 hours on standard hardware. The pipeline also provides comprehensive ML benchmarks across seven model architectures and four clinical prediction tasks, establishing reproducible baselines for future research.

\textbf{Broader Impact.} By making the groundbreaking CRITICAL dataset immediately accessible, CRISP democratizes multi-institutional healthcare AI research. The pipeline transforms months of manual data curation into a ready-to-deploy solution, lowering the barrier from requiring specialized data engineering expertise to simply executing pre-configured scripts. This accessibility significantly lowers the barrier to entry, enabling researchers to focus on developing and testing innovative models rather than wrestling with data preprocessing challenges. CRISP's comprehensive infrastructure enables the research community to quickly leverage CRITICAL's unprecedented scale and diversity, accelerating progress toward robust, generalizable clinical AI systems. The combination of systematic data processing, concept harmonization, and reproducible benchmarks establishes a foundation for collaborative advancement in cross-institutional healthcare ML.

\textbf{Limitations and Future Work.} While current vocabulary mapping prioritizes four critical tables that exhibit the most complex heterogeneity and are essential for clinical prediction, future releases will progressively extend mapping coverage to all tables requiring harmonization. The pipeline currently employs empirically-derived parameters—such as 99th percentile outlier thresholds and 2-hour visit merging windows—that provide robust general-purpose processing but may not be optimal for specific clinical tasks. Future work may develop task-adaptive parameter selection, leverage CRITICAL's unique longitudinal coverage spanning pre-ICU, ICU, and post-ICU periods to explore novel prediction tasks with high clinical value, and investigate advanced feature utilization strategies for the dataset's extensive concepts.


\newpage
\bibliography{references}

\newpage

\appendix

\section{Dataset Demographics and Volume Statistics}

This appendix provides detailed demographic and volume statistics that demonstrate the multi-institutional heterogeneity CRISP was designed to address. Table~\ref{tab:appendix_cohort} presents the comprehensive demographic breakdown of 371,365 patients across four CTSA sites, showcasing the substantial diversity that validates CRISP's generalizability across different patient populations. The demographic representation across race, ethnicity, and age groups exemplifies the health equity research opportunities enabled by CRISP's systematic vocabulary harmonization, which ensures consistent feature representation across diverse institutional practices and coding standards.

Table~\ref{tab:appendix_volumes} illustrates the scale of vocabulary harmonization challenges addressed by CRISP—processing 1.95 billion records across 17 OMOP CDM tables with systematic standardization. The MEASUREMENT table alone contains 1.4 billion records, representing the largest single source of clinical data and exemplifying the computational challenges that CRISP efficiently handles through parallel processing and intelligent chunking. These volumes, combined with the extensive temporal coverage shown in Table~\ref{tab:appendix_temporal} (median observation period of 3.11 years, with 65.6\% of patients having multi-year longitudinal data), demonstrate CRISP's ability to transform massive, heterogeneous multi-institutional data into ML-ready datasets within approximately 20 hours on standard hardware, democratizing access to large-scale critical care data for the broader research community.

\begin{table*}[!ht]
\centering
\caption{Demographic characteristics of 371,365 patients in the CRITICAL dataset across four CTSA sites, demonstrating the multi-institutional diversity in race, ethnicity, gender, and age distributions}
\label{tab:appendix_cohort}
\footnotesize
\begin{tabular}{p{2.9cm}p{3.5cm}rrrr}
\toprule
\rowcolor{headerblue!20}
\multicolumn{2}{c}{\textbf{Variable}} & \multicolumn{2}{c}{\textbf{Gender}} & \multicolumn{2}{c}{\textbf{Total}} \\
\cmidrule(lr){3-4} \cmidrule(lr){5-6}
\multicolumn{2}{c}{} & \textbf{Male} & \textbf{Female} & \textbf{N} & \textbf{\%} \\
\midrule

\rowcolor{gray!10}
\textbf{Race} & & & & & \\
& Asian$^a$ & 6,750 {\textcolor{gray}{\scriptsize
 (55.17\%)}} & 5,483 {\textcolor{gray}{\scriptsize (44.83\%)}} & 12,233 & 3.29\% \\
& Black/African American & 39,142 {\textcolor{gray}{\scriptsize (52.62\%)}} & 35,237 {\textcolor{gray}{\scriptsize (47.38\%)}} & 74,382 & 20.03\% \\
& Native American & 338 {\textcolor{gray}{\scriptsize (55.23\%)}} & 274 {\textcolor{gray}{\scriptsize (44.77\%)}} & 612 & 0.16\% \\
& Pacific Islander$^b$ & 426 {\textcolor{gray}{\scriptsize (54.27\%)}} & 359 {\textcolor{gray}{\scriptsize (45.73\%)}} & 785 & 0.21\% \\
& White & 141,993 {\textcolor{gray}{\scriptsize (56.72\%)}} & 108,327 {\textcolor{gray}{\scriptsize (43.28\%)}} & 250,328 & 67.41\% \\
& Multiple Race & 1,612 {\textcolor{gray}{\scriptsize (59.01\%)}} & 1,120 {\textcolor{gray}{\scriptsize (41.00\%)}} & 2,732 & 0.74\% \\
& Unknown & 9,217 {\textcolor{gray}{\scriptsize (56.43\%)}} & 7,116 {\textcolor{gray}{\scriptsize (43.57\%)}} & 16,337 & 4.40\% \\
& Other/Refused & 8,121 {\textcolor{gray}{\scriptsize (58.19\%)}} & 5,830 {\textcolor{gray}{\scriptsize (41.81\%)}} & 13,956 & 3.76\% \\

\midrule
\rowcolor{gray!10}
\textbf{Ethnicity} & & & & & \\
& Hispanic/Latino & 12,065 {\textcolor{gray}{\scriptsize (56.76\%)}} & 9,195 {\textcolor{gray}{\scriptsize (43.24\%)}} & 21,260 & 5.72\% \\
& Not Hispanic/Latino & 185,417 {\textcolor{gray}{\scriptsize (55.78\%)}} & 146,949 {\textcolor{gray}{\scriptsize (44.22\%)}} & 332,376 & 89.50\% \\
& Unknown & 10,117 {\textcolor{gray}{\scriptsize (57.07\%)}} & 7,602 {\textcolor{gray}{\scriptsize (42.93\%)}} & 17,729 & 4.77\% \\

\midrule
\rowcolor{gray!10}
\textbf{Age at First Visit$^c$} & & & & & \\
& $<$18 & 40,591 {\textcolor{gray}{\scriptsize (55.34\%)}} & 32,753 {\textcolor{gray}{\scriptsize (44.66\%)}} & 73,352 & 19.75\% \\
& 18-30 & 13,775 {\textcolor{gray}{\scriptsize (53.60\%)}} & 11,926 {\textcolor{gray}{\scriptsize (46.40\%)}} & 25,703 & 6.92\% \\
& 31-50 & 39,876 {\textcolor{gray}{\scriptsize (56.62\%)}} & 30,545 {\textcolor{gray}{\scriptsize (43.38\%)}} & 70,421 & 18.96\% \\
& 51-70 & 79,063 {\textcolor{gray}{\scriptsize (58.41\%)}} & 56,306 {\textcolor{gray}{\scriptsize (41.59\%)}} & 135,372 & 36.45\% \\
& $>$70 & 34,294 {\textcolor{gray}{\scriptsize (51.56\%)}} & 32,216 {\textcolor{gray}{\scriptsize (48.44\%)}} & 66,517 & 17.91\% \\

\midrule
\rowcolor{gray!10}
\textbf{Visit Type$^d$} & & & & & \\
& Outpatient & 7,542,966 {\textcolor{gray}{\scriptsize (51.50\%)}} & 7,105,372 {\textcolor{gray}{\scriptsize (48.50\%)}} & 14,648,554 & 52.37\% \\
& Inpatient (non-ICU) & 3,335,192 {\textcolor{gray}{\scriptsize (53.02\%)}} & 2,954,536 {\textcolor{gray}{\scriptsize (46.98\%)}} & 6,289,940 & 22.49\% \\
& ICU & 506,178 {\textcolor{gray}{\scriptsize (57.08\%)}} & 380,685 {\textcolor{gray}{\scriptsize (42.92\%)}} & 886,896 & 3.17\% \\
& Emergency & 385,437 {\textcolor{gray}{\scriptsize (51.04\%)}} & 369,800 {\textcolor{gray}{\scriptsize (48.96\%)}} & 755,243 & 2.70\% \\
& Other & 2,868,848 {\textcolor{gray}{\scriptsize (53.21\%)}} & 2,522,983 {\textcolor{gray}{\scriptsize (46.79\%)}} & 5,391,831 & 19.28\% \\

\midrule
\rowcolor{gray!10}
\textbf{Mortality$^e$} & & & & & \\
& Alive & 162,062 {\textcolor{gray}{\scriptsize (56.02\%)}} & 127,243 {\textcolor{gray}{\scriptsize (43.98\%)}} & 289,321 & 77.91\% \\
& Deceased & 45,537 {\textcolor{gray}{\scriptsize (55.50\%)}} & 36,503 {\textcolor{gray}{\scriptsize (44.50\%)}} & 82,044 & 22.09\% \\

\midrule
\midrule
\rowcolor{headerblue!20}
\textbf{Total}$^f$ & & 207,599 {\textcolor{gray}{\scriptsize (55.91\%)}} & 163,746 {\textcolor{gray}{\scriptsize (44.09\%)}} & 371,365 & 100.00\% \\
\bottomrule
\end{tabular}

\vspace{0.2cm}
\footnotesize
\textit{Notes:}\\
$^a$ Asian includes: Asian (11,227), Asian Indian (506), Korean (206), Chinese (100), Japanese (83), Vietnamese (49), Filipino (25), Thai (24), Cambodian (13)\\
$^b$ Pacific Islander includes: Native Hawaiian (557), Native Hawaiian or Other Pacific Islander (228)\\
$^c$ Age calculated as (Earliest Visit Date - Birth Date) / 365.25\\
$^d$ Visit Type based on 27,972,464 visit details from VISIT\_DETAIL table. ICU identified by concept IDs 32037, 581379\\
$^e$ Mortality status reflects patient vital status at last recorded encounter in the dataset\\
$^f$ Total includes 20 additional patients with unknown or missing gender concept IDs
\end{table*}

\begin{table*}[!ht]
\centering
\caption{Data volume distribution across 17 OMOP CDM tables in the CRITICAL dataset, totaling 1.95 billion records, with average records per patient demonstrating the comprehensive clinical coverage}
\label{tab:appendix_volumes}
\footnotesize
\begin{tabular}{@{}l r r r r@{}}
\toprule
\rowcolor{headerblue!20}
\textbf{Table Name} & \textbf{Row Count} & \textbf{Size (GB)} & \textbf{\% of Total} & \textbf{Per Patient} \\
\midrule
MEASUREMENT & 1,403,627,644 & 194.00 & 72.08\% & 3,779.1 \\
\rowcolor{gray!10}
OBSERVATION & 174,355,400 & 21.06 & 8.95\% & 469.4 \\
DRUG\_EXPOSURE & 160,361,417 & 27.18 & 8.24\% & 431.7 \\
\rowcolor{gray!10}
CONDITION\_OCCURRENCE & 138,749,128 & 17.63 & 7.13\% & 373.6 \\
VISIT\_OCCURRENCE & 38,000,960 & 5.29 & 1.95\% & 102.3 \\
\rowcolor{gray!10}
CONDITION\_ERA & 35,921,008 & 2.73 & 1.85\% & 96.7 \\
PROCEDURE\_OCCURRENCE & 31,905,907 & 3.71 & 1.64\% & 85.9 \\
\rowcolor{gray!10}
VISIT\_DETAIL & 27,972,464 & 4.44 & 1.44\% & 75.3 \\
DRUG\_ERA & 24,322,578 & 1.88 & 1.25\% & 65.5 \\
\rowcolor{gray!10}
DEVICE\_EXPOSURE & 5,212,843 & 0.66 & 0.27\% & 14.0 \\
LOCATION & 4,875,096 & 0.10 & 0.25\% & 13.1 \\
\rowcolor{gray!10}
SPECIMEN & 2,123,886 & 0.17 & 0.11\% & 5.7 \\
PROVIDER & 623,239 & 0.02 & 0.03\% & 1.7 \\
\rowcolor{gray!10}
PERSON & 371,365 & 0.04 & 0.02\% & 1.0 \\
OBSERVATION\_PERIOD & 244,350 & 0.01 & 0.01\% & 0.7 \\
\rowcolor{gray!10}
DEATH & 82,064 & 0.004 & 0.004\% & 0.2 \\
CARE\_SITE & 5,966 & 0.0002 & 0.0003\% & / \\
\midrule
\rowcolor{headerblue!20}
\textbf{TOTAL} & \textbf{1,947,180,421} & \textbf{278.97} & \textbf{100.00\%} & \textbf{5,242.2} \\
\bottomrule
\end{tabular}
\end{table*}

\begin{table*}[!ht]
\centering
\caption{Temporal characteristics of the CRITICAL dataset showing extensive longitudinal coverage with median observation period of 3.11 years and 65.6\% of patients having multi-year data}
\label{tab:appendix_temporal}
\footnotesize
\begin{tabular}{lr}
\toprule
\rowcolor{headerblue!20}
\textbf{Temporal Characteristic} & \textbf{Value} \\
\midrule
\multicolumn{2}{c}{\textbf{Time Span Statistics}} \\
\rowcolor{gray!10}
Mean observation period & 1,983.6 days (5.43 years) \\
Median observation period & 1,137.0 days (3.11 years) \\
\rowcolor{gray!10}
Standard deviation & 2,168.9 days (5.94 years) \\
Minimum time span & 0 days \\
\rowcolor{gray!10}
Maximum time span & 11,631 days (31.8 years) \\
25th percentile & 94.0 days (0.26 years) \\
\rowcolor{gray!10}
75th percentile & 3,463.0 days (9.49 years) \\
\midrule
\multicolumn{2}{c}{\textbf{Visit Frequency}} \\
\rowcolor{gray!10}
Mean visits per patient & 102.3 \\
Median visits per patient & 24 \\
\midrule
\multicolumn{2}{c}{\textbf{Patient Records Time Span Distribution}$^a$} \\
\rowcolor{gray!10}
Single visit (0 days) & 2,823 (0.8\%) \\
Under 1 year & 127,692 (34.4\%) \\
\rowcolor{gray!10}
1-5 years & 100,125 (27.0\%) \\
5-10 years & 56,139 (15.1\%) \\
\rowcolor{gray!10}
10-15 years & 46,182 (12.4\%) \\
15-20 years & 40,356 (10.9\%) \\
\rowcolor{gray!10}
Over 20 years & 871 (0.2\%) \\
\midrule
\rowcolor{headerblue!20}
\textbf{Total Patients} & \textbf{371,365 (100.0\%)} \\
\bottomrule
\end{tabular}
\vspace{0.2cm}
\footnotesize

\textit{Notes:}

$^a$ Time span calculated as the difference between the last and first visit times in the VISIT\_OCCURRENCE table for each patient. \\

\end{table*}

\section{Vocabulary Distribution Analysis}
\label{apd:second}

This section visualizes the vocabulary heterogeneity challenges that necessitate CRISP's cross-vocabulary mapping (Stage 3). The fragmentation of clinical concepts across multiple vocabularies creates sparse feature matrices that impede effective machine learning, making systematic harmonization to unified SNOMED standards essential.

\begin{figure*}[!ht]
\centering
\includegraphics[width=0.72\textwidth]{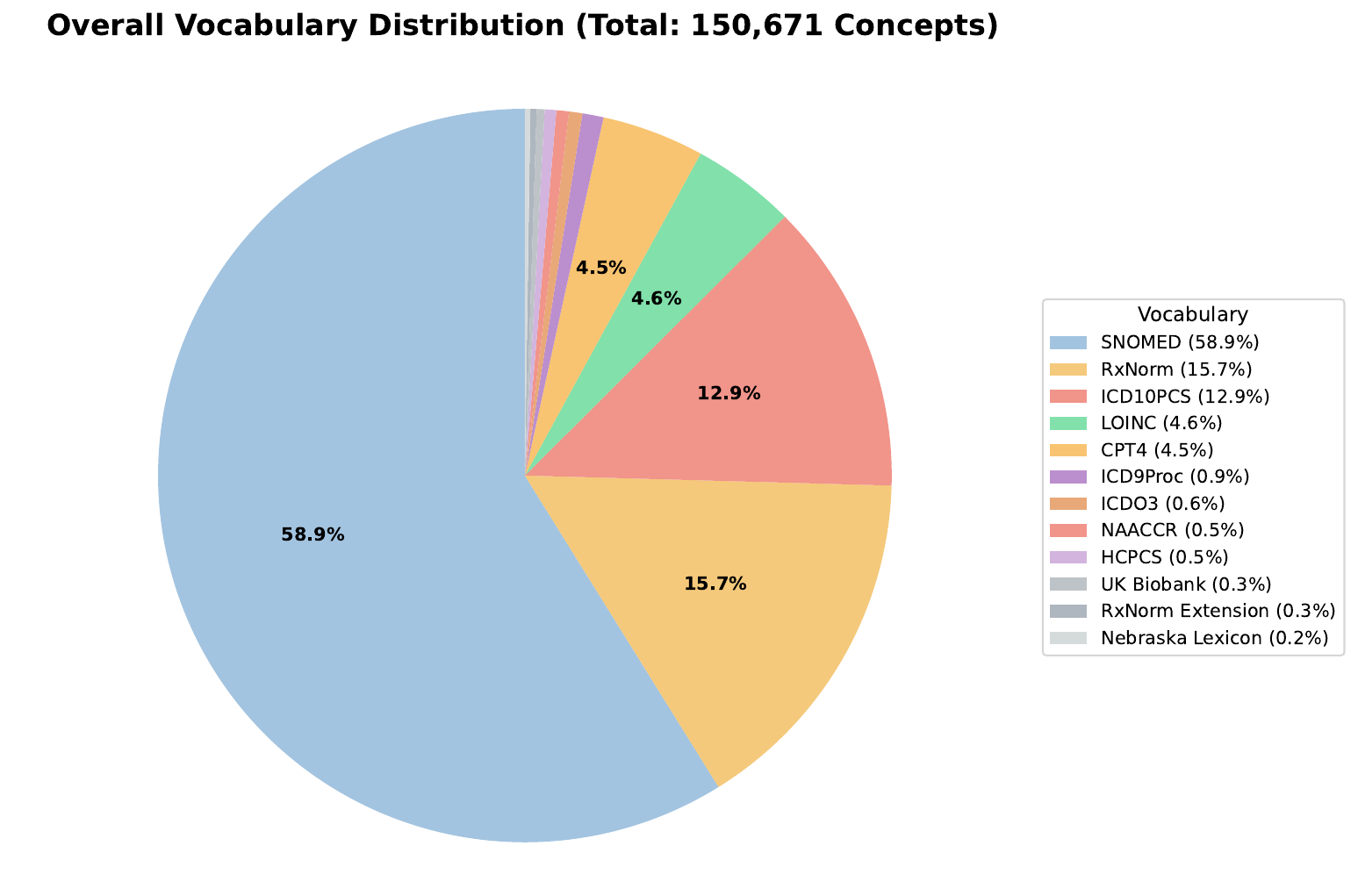}
\caption{Overall vocabulary distribution across 150,671 unique concepts in the CRITICAL dataset. SNOMED represents 58.0\% (87,453 concepts), followed by RxNorm (15.7\%), ICD10PCS (12.9\%), illustrating the heterogeneity challenge addressed by CRISP's mapping stage.}
\label{fig:vocab_pie_appendix}
\end{figure*}

\begin{figure*}[!ht]
\centering
\includegraphics[width=0.8\textwidth]{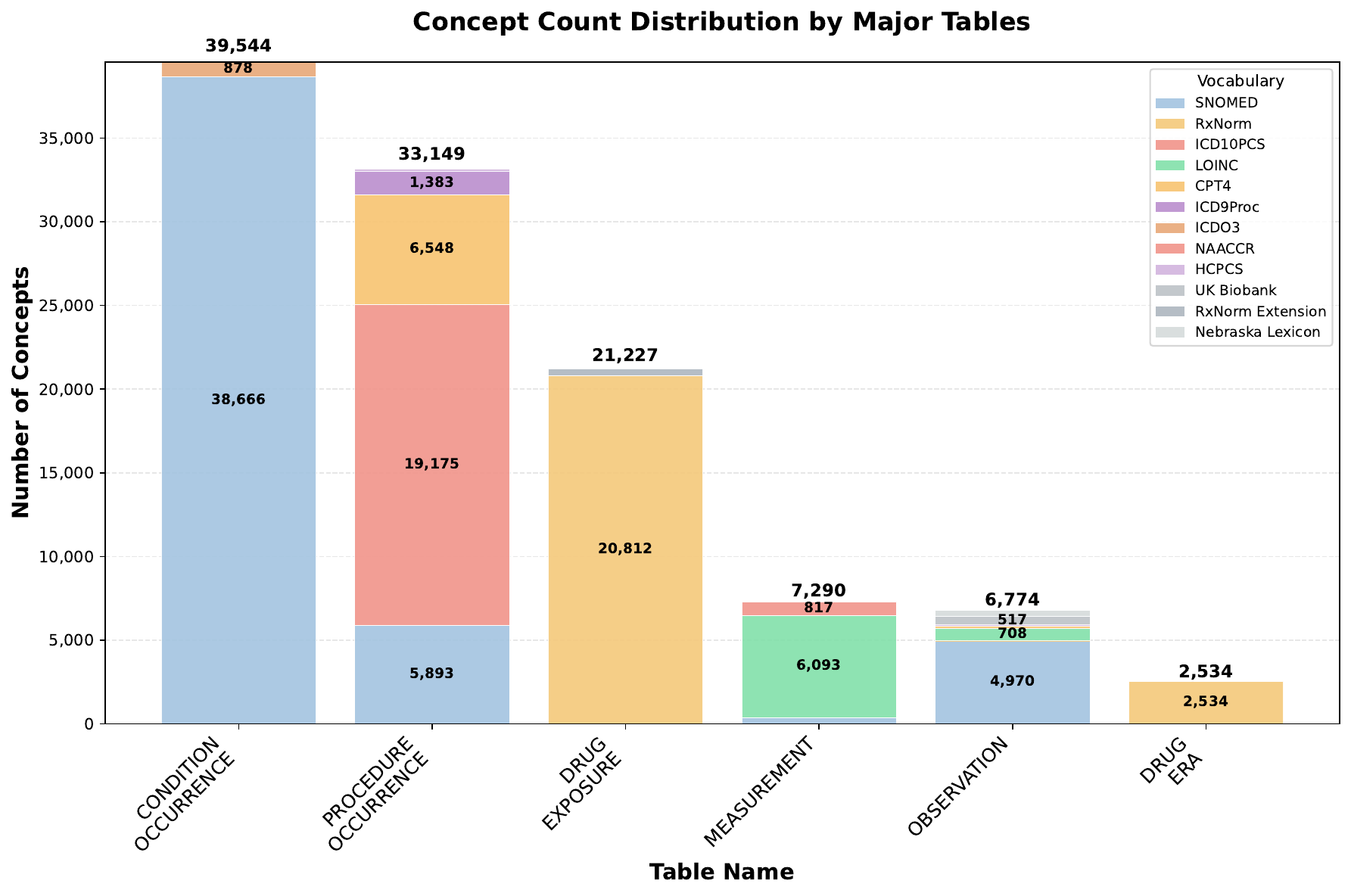}
\caption{Absolute concept count distribution across major OMOP tables. CONDITION\_OCCURRENCE exhibits the highest vocabulary diversity (39,544 unique concepts), followed by PROCEDURE\_OCCURRENCE (33,149 concepts), while specialized tables like DRUG\_ERA show focused vocabularies (2,534 concepts), demonstrating domain-specific vocabulary patterns.}
\label{fig:vocab_stacked_appendix}
\end{figure*}

Figure \ref{fig:vocab_pie_appendix} shows the distribution of unique concepts across vocabularies, with SNOMED representing the majority while substantial portions use RxNorm, ICD10PCS, and other specialized terminologies. Figure \ref{fig:vocab_stacked_appendix} reveals the absolute concept counts within major OMOP tables, demonstrating how vocabulary usage varies significantly across clinical domains—from CONDITION\_OCCURRENCE's 39,544 concepts to DRUG\_ERA's focused 2,534 concepts. These visualizations underscore the complexity of harmonizing diverse medical terminologies across multi-institutional datasets.

\section{Deep Learning Model Architectures}
\label{apd:architectures}

This section presents the detailed architectures of two deep learning models used in our benchmark experiments. Both the LSTM and TCN models employ a hybrid architecture that integrates static patient features with temporal clinical measurements, enabling comprehensive representation learning from the multi-modal CRITICAL dataset. 

\begin{figure*}[!ht]
\centering
\includegraphics[width=0.8\textwidth]{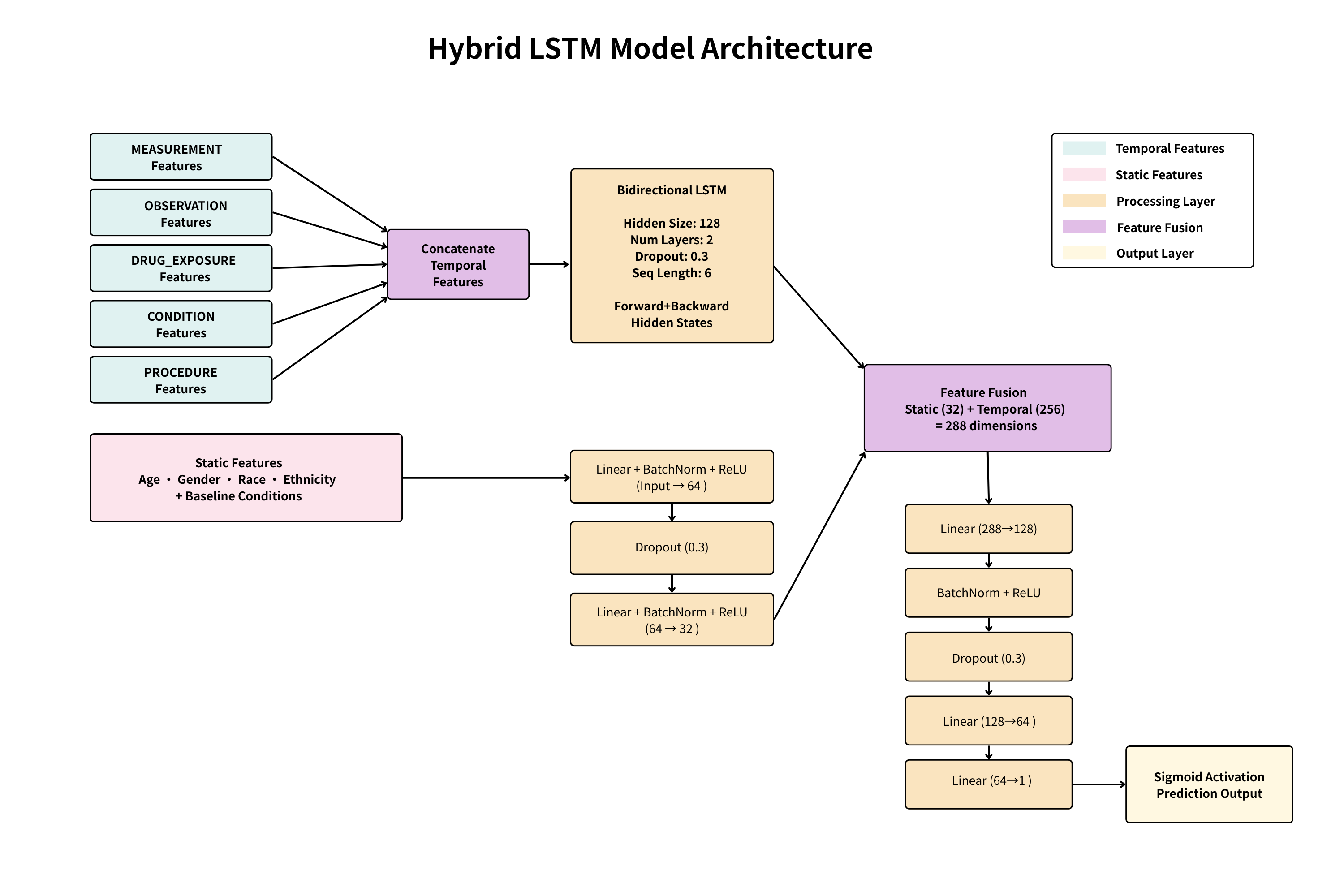}
\caption{LSTM hybrid architecture with bidirectional LSTM layers for temporal features and separate encoding for static patient features.}
\label{fig:lstm_architecture}
\end{figure*}

\begin{figure*}[!ht]
\centering
\includegraphics[width=0.8\textwidth]{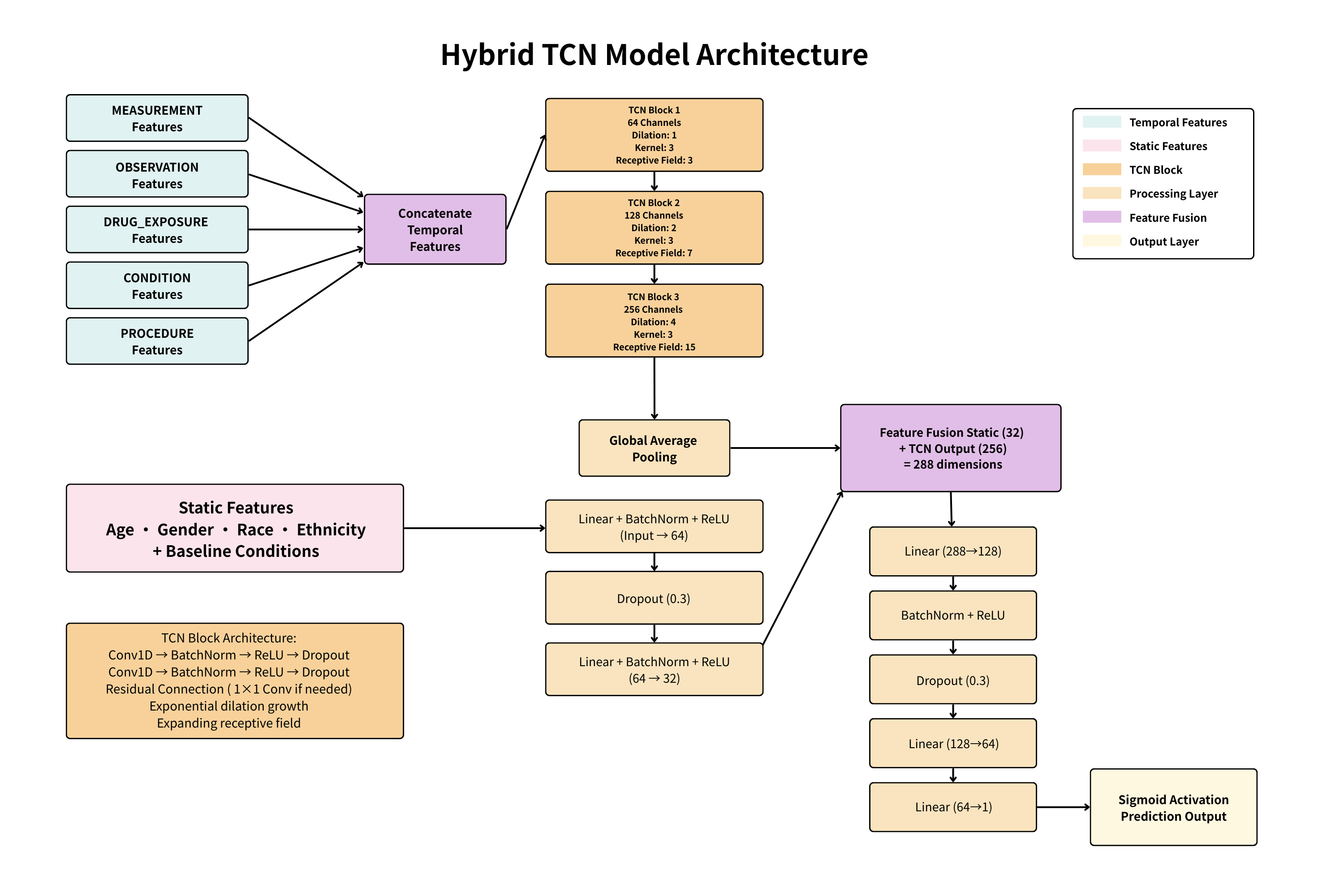}
\caption{TCN hybrid architecture with dilated causal convolutions (dilation factors: 1, 2, 4) for temporal features and parallel processing for static features. The temporal features composition is identical to the LSTM architecture.}
\label{fig:tcn_architecture}
\end{figure*}

\end{document}